%% file: main.tex
\documentclass[conference,a4paper]{IEEEtran}
\IEEEoverridecommandlockouts

\usepackage{amsmath}
\usepackage{amssymb}
\usepackage{amsfonts}
\usepackage{mathtools}

\usepackage{booktabs}       
\usepackage{multirow}       
\usepackage{array}          
\usepackage{graphicx}       
\usepackage{adjustbox}      

\usepackage{tikz}           
\usepackage{float}          
\usepackage{subfigure}      
\usepackage{caption}

\usepackage{algorithm}
\usepackage{algorithmic}

\usepackage{xcolor}         
\usepackage{url}            
\usepackage{enumitem}       
\usepackage{bbm}            
\usepackage{bm}             
\usepackage{dsfont}         
\usepackage{hyperref}       

\usepackage{titlesec}       
\usepackage{cleveref}       
 \usepackage{amssymb}
 
\def\BibTeX{{\rm B\kern-.05em{\sc i\kern-.025em b}\kern-.08em
    T\kern-.1667em\lower.7ex\hbox{E}\kern-.125emX}}
\begin{document}

\title{DPL: Spatial-Conditioned Diffusion Prototype Enhancement for One-Shot Medical Segmentation
}

\author{\IEEEauthorblockN{1\textsuperscript{st} Ziyuan Gao}
\IEEEauthorblockA{
\textit{University College London}\\
London, UK \\
ucbqzg5@ucl.ac.uk \\
ORCID$:$ 0009-0005-8092-5576}
\and
\IEEEauthorblockN{2\textsuperscript{nd} Philippe Morel}
\IEEEauthorblockA{
\textit{University College London}\\
London, UK \\
p.morel@ucl.ac.uk \\
ORCID$:$ 0000-0002-9847-3409
}
}

\IEEEoverridecommandlockouts
   \IEEEpubid{\makebox[\columnwidth]{979-8-3315-8654-6/25/\$31.00~\copyright2025 IEEE \hfill}
   \hspace{\columnsep}\makebox[\columnwidth]{ }}
   
\maketitle
\IEEEpubidadjcol

\begin{abstract}
One-shot medical image segmentation faces fundamental challenges in prototype representation due to limited annotated data and significant anatomical variability across patients. Traditional prototype-based methods rely on deterministic averaging of support features, creating brittle representations that fail to capture intra-class diversity essential for robust generalization. This work introduces Diffusion Prototype Learning (DPL), a novel framework that reformulates prototype construction through diffusion-based feature space exploration. DPL models one-shot prototypes as learnable probability distributions, enabling controlled generation of diverse yet semantically coherent prototype variants from minimal labeled data. The framework operates through three core innovations: (1) a diffusion-based prototype enhancement module that transforms single support prototypes into diverse variant sets via forward-reverse diffusion processes, (2) a spatial-aware conditioning mechanism that leverages geometric properties derived from prototype feature statistics, and (3) a conservative fusion strategy that preserves prototype fidelity while maximizing representational diversity. 
DPL ensures training-inference consistency by using the same diffusion enhancement and fusion pipeline in both phases. This process generates enhanced prototypes that serve as the final representations for similarity calculations, while the diffusion process itself acts as a regularizer.
Extensive experiments on abdominal MRI and CT datasets demonstrate significant improvements respectively, establishing new state-of-the-art performance in one-shot medical image segmentation.
\end{abstract}

\begin{IEEEkeywords}
One-shot segmentation, Diffusion models, Medical image analysis
\end{IEEEkeywords}

\input{1.introduction}

\input{2.relatedwork}

\input{3.methodology}
\input{4.experiment}

\bibliographystyle{IEEEtran}  
\bibliography{references}

\end{document}

%% file: 1.introduction.tex
\section{Introduction}

\begin{figure*}[h]
    \centering
    \includegraphics[width=0.9\textwidth]{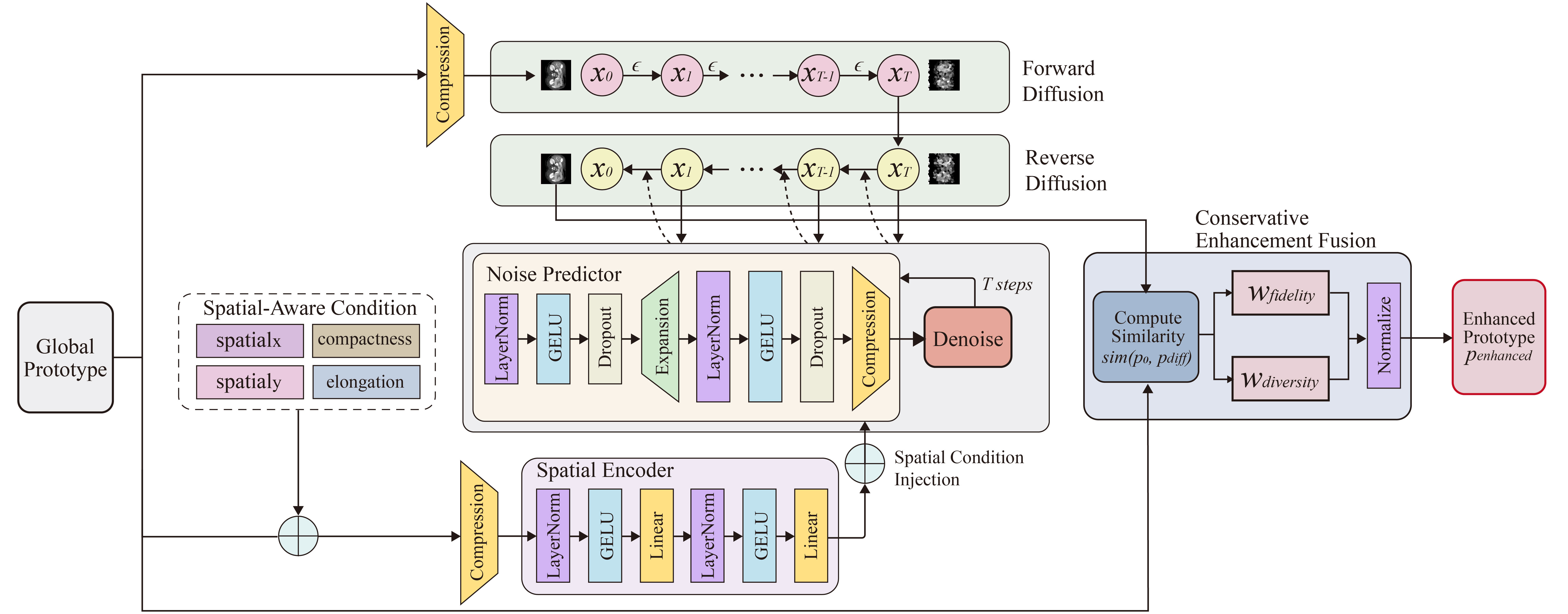}
    \caption{Architecture of one-shot prototype diffusion enhancer}
    \label{fig:architecture}
\end{figure*}

One-shot learning has emerged as a critical paradigm for medical image segmentation, addressing the fundamental challenge of limited annotated data in clinical applications~\cite{ref1,ref2}. Unlike natural image domains where extensive labeled datasets are readily available, medical imaging suffers from inherent data scarcity due to privacy constraints, annotation costs, and expert knowledge requirements~\cite{ref3}. This scarcity is pronounced in medical segmentation tasks, where pixel-level annotations demand significant clinical expertise.

Traditional one-shot segmentation approaches rely on prototype-based learning, where support examples are condensed into representative prototypes through simple averaging operations~\cite{ref4,ref5}. While computationally efficient, this deterministic approach fundamentally limits representational capacity, as it collapses diverse morphological variations into single point estimates. Moreover, medical images exhibit substantial intra-class variability due to patient demographics, imaging protocols, pathological conditions, and anatomical variations~\cite{ref6}. Such diversity cannot be adequately captured through deterministic averaging, leading to brittle prototypes. Consequently, these prototypes fail to generalize across the diverse imaging patterns encountered in clinical practice.

Recent advances in diffusion models have demonstrated remarkable success in generative tasks, offering principled approaches to modeling complex data distributions~\cite{ref7}. These models excel at capturing underlying data manifolds and generating diverse yet coherent samples, making them particularly suited for addressing the prototype diversity challenge in one-shot learning. However, direct application of diffusion models to one-shot segmentation presents unique challenges, including computational efficiency, spatial coherence preservation, and integration with existing prototype-based architectures.

This work introduces Diffusion Prototype Learning (DPL), a novel framework that reformulates prototype construction through diffusion-based feature space exploration. Unlike traditional deterministic approaches, DPL models prototypes as learnable probability distributions, enabling controlled generation of diverse prototype variants while preserving semantic consistency. The key insight lies in leveraging diffusion processes not for data generation, but for prototype enhancement through systematic feature space exploration. To achieve this, DPL trains a diffusion enhancer network with spatial conditioning that generates diverse prototype variants from single support examples through controlled noise injection and denoising. This approach effectively captures medical imaging variability without requiring additional annotated data.

The main contributions of this work are threefold: 

(1) A diffusion-based prototype enhancement framework that transforms single support prototypes into diverse, semantically coherent variants. The framework uses forward-reverse diffusion processes to address the fundamental limitation of deterministic averaging in one-shot learning. (2) A spatial-aware conditioning mechanism is introduced to guide the diffusion process with geometric constraints. This ensures generated prototype variants maintain anatomical validity. (3) A conservative fusion strategy that preserves prototype fidelity while incorporating diffusion-generated diversity ensuring training-inference consistency through identical enhancement pipelines across both phases. Extensive experiments have been conducted on abdominal MRI and abdominal CT datasets to evaluate the proposed DPL framework. The results demonstrate significant improvements over state-of-the-art methods. This establishes new performance benchmarks in one-shot medical image segmentation.

%% file: 2.relatedwork.tex
\section{Related Work}

\subsection{One-Shot Medical Image Segmentation}

One-shot segmentation in medical imaging presents unique challenges due to domain-specific constraints and significant anatomical variability. Medical image analysis requires methods that can handle substantial morphological variations across patients, imaging protocols, and pathological conditions. Ouyang et al.~\cite{ref12} developed self-supervision mechanisms that iteratively refine prototypes through pseudo-label generation in medical contexts. Hansen et al.~\cite{ref13} explored anomaly detection-inspired one-shot medical segmentation through self-supervision with supervoxels. While general one-shot segmentation methods like PANet~\cite{ref5}, ALPNet~\cite{ref10}, and DSPNet~\cite{ref11} have shown promise in natural image domains, their direct application to medical imaging reveals fundamental limitations. These approaches continue to rely on deterministic prototype extraction methods that fail to capture the full spectrum of anatomical variations present in clinical data, highlighting the need for specialized approaches that can effectively model the inherent diversity in medical imaging.

\subsection{Diffusion Models in Medical Image Analysis}

Diffusion models have demonstrated significant potential in medical image analysis, particularly for addressing anatomical variability and data scarcity challenges. Rombach et al.~\cite{ref14} developed latent diffusion models that have been adapted for medical image synthesis, enabling the generation of diverse anatomical variations while preserving pathological realism. Ho et al.~\cite{ref7} introduced denoising diffusion probabilistic models (DDPMs) that have shown promise in medical data augmentation by generating synthetic medical images that capture morphological diversity. Song et al.~\cite{ref8} extended this framework through score-based generative modeling, which has been effective in handling the complex probability distributions inherent in medical imaging data. Recent works have begun exploring diffusion models for medical image enhancement and restoration, though their application to prototype enhancement in one-shot medical segmentation remains largely unexplored, representing an opportunity for improving segmentation performance in data-limited clinical scenarios.

\subsection{Prototype Enhancement in One-Shot Learning}

Traditional prototype enhancement strategies in one-shot learning remain fundamentally limited by their reliance on deterministic operations. Snell et al.~\cite{ref4} established prototypical networks employing simple averaging operations that cannot adequately capture the diversity required for robust one-shot performance in challenging domains such as medical imaging. Finn et al.~\cite{ref15} developed gradient-based meta-learning approaches that, while effective for rapid adaptation, still rely on deterministic prototype representations.

%% file: 3.methodology.tex
\section{Methodology}

\subsection{Initial Prototype Extraction}

In medical image segmentation, anatomical structures exhibit significant morphological variability across patients and imaging conditions~\cite{ref7,ref8}, making prototype based one-shot learning particularly challenging~\cite{ref4}. Traditional prototype extraction methods rely on simple averaging of support features~\cite{ref5}, which often fails to capture the inherent diversity of organ shapes, tissue appearances, and pathological variations present in medical data~\cite{ref6}. DPL addresses this fundamental limitation by transforming prototype extraction from a deterministic averaging operation into a probabilistic generation process that can systematically explore semantically valid neighborhoods in feature space.

Given support features $\mathbf{X}_s \in \mathbb{R}^{C \times H \times W}$ and corresponding masks $\mathbf{Y}_s$, initial prototypes are extracted through masked averaging:

\begin{equation}
\mathbf{p}_0 = \frac{\sum_{h,w} \mathbf{X}_s(:,h,w) \cdot \mathbf{Y}_s(h,w)}{\sum_{h,w} \mathbf{Y}_s(h,w) + \epsilon}
\end{equation}

This initial extraction uses a conventional approach to create a deterministic anchor point, the prototype $p_0 \in \mathbb{R}^{256}$. The diffusion process then treats this anchor point as a sample from an underlying prototype distribution, which allows for probabilistic enhancement.


\begin{figure}
    \centering
    \includegraphics[width=\columnwidth]{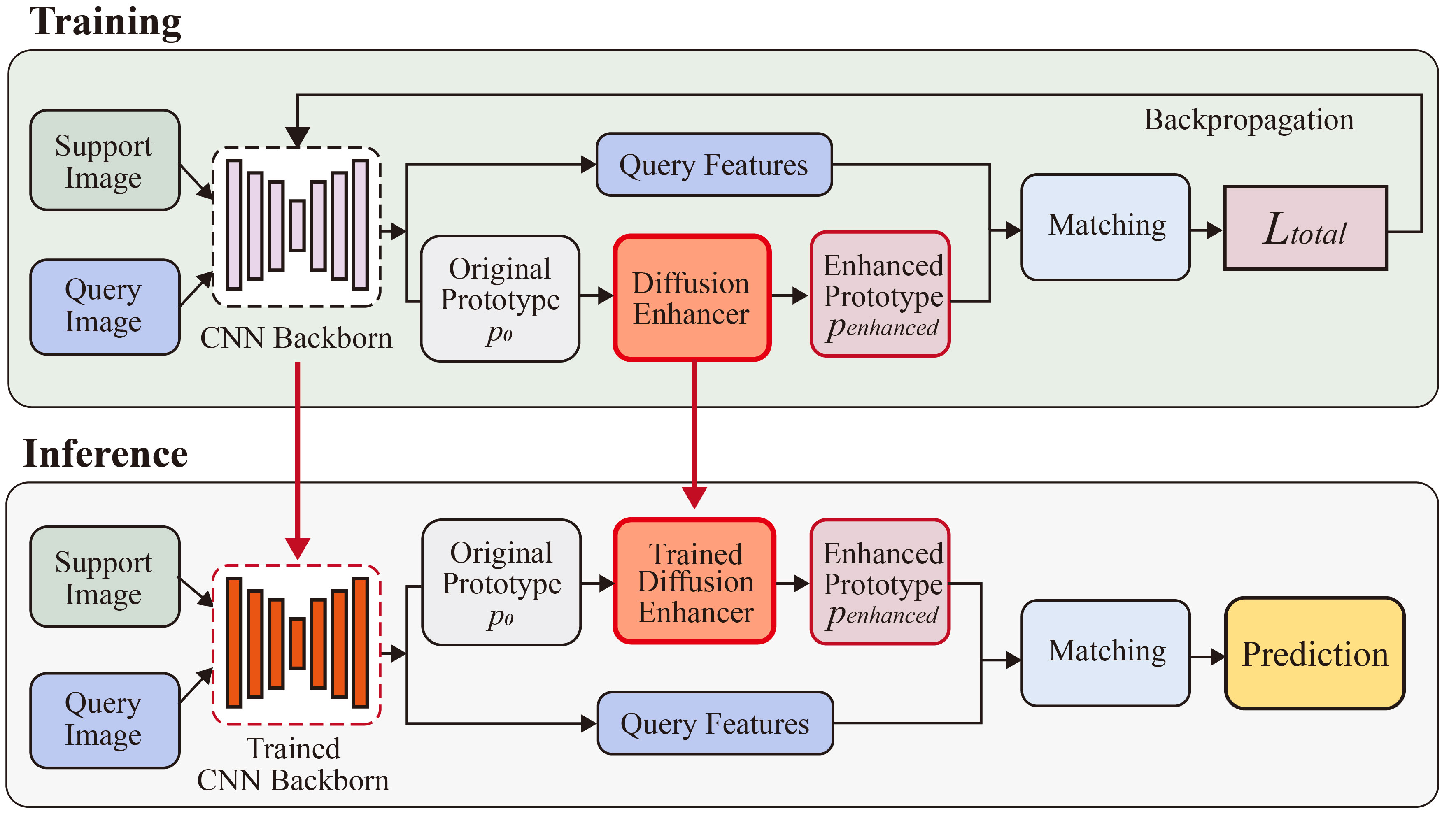}
    \caption{Architecture of training and inference phases for one-shot learning.}
    \label{fig:overall}
\end{figure}

\subsection{One-Shot Prototype Diffusion Enhancer}

Figure~\ref{fig:architecture} and ~\ref{fig:overall} present the comprehensive architecture of the proposed one-shot prototype diffusion enhancer framework. The following sections provide detailed description of its key components and mechanisms.

\noindent\textbf{Forward Diffusion for Feature Augmentation.}
The forward diffusion process serves as a controlled noise injection mechanism to create enhanced prototype variants from limited support features. Rather than training a full generative model, this approach uses diffusion as a principled way to perturb prototypes and generate variations for one-shot learning.

Given an initial prototype $\mathbf{p}_0 \in \mathbb{R}^D$ extracted from support features, the forward process adds Gaussian noise at a selected timestep $t \sim \text{Uniform}(1, T)$:

\begin{equation}
\mathbf{p}_t = \sqrt{\bar{\alpha}_t} \mathbf{p}_0 + \sqrt{1-\bar{\alpha}_t} \boldsymbol{\epsilon}
\end{equation}

where $\boldsymbol{\epsilon} \sim \mathcal{N}(0, \mathbf{I})$ and $\bar{\alpha}_t = \prod_{i=1}^{t} \alpha_i$ with $\alpha_t = 1 - \beta_t$. The noise schedule follows a cosine beta schedule with $\beta_t$ clipped to $[0.001, 0.1]$ over $T=20$ timesteps. This approach enables the generation of diverse prototype variants without requiring extensive training data, addressing the fundamental challenge of prototype scarcity in one-shot segmentation.

\noindent\textbf{Reverse Diffusion for Prototype Refinement.}
The reverse diffusion process reconstructs enhanced prototypes from the noisy variants generated in the forward pass. This refinement is achieved through a learned denoising procedure. The process serves as a feature refinement mechanism that preserves semantic information, while introducing controlled variations that are beneficial for one-shot segmentation.

Given a noisy prototype $\mathbf{p}_t$ at timestep $t$, the reverse process iteratively denoises through timesteps $t, t-1, \ldots, 1, 0$ using a noise prediction network $\epsilon_\theta$:

\begin{equation}
\mathbf{p}_{t-1} = \frac{\sqrt{\alpha_{t-1}} \beta_t}{1-\bar{\alpha}_t} \hat{\mathbf{p}}_0 + \frac{\sqrt{\alpha_t}(1-\bar{\alpha}_{t-1})}{1-\bar{\alpha}_t} \mathbf{p}_t + \sigma_t \mathbf{z}_t
\end{equation}

where $\mathbf{z}_t \sim \mathcal{N}(0, \mathbf{I})$ represents additional stochastic noise.

\noindent\textbf{Spatial-Aware Condition Encoding and Injection.}
Medical image segmentation demands spatial coherence due to fixed anatomical relationships and geometric constraints~\cite{ref7}. To ensure generated prototype variants preserve these spatial dependencies, this work incorporates a spatial conditioning mechanism that guides the diffusion process with geometric information~\cite{ref16}.

The spatial conditioning integrates prototype features with geometric constraints derived from feature statistics through a dedicated encoder:
\begin{equation}
\mathbf{c}_{spatial} = f_{enc}([\mathbf{p}_0; \mathbf{g}_{geom}])
\end{equation}
where $\mathbf{g}_{geom} \in \mathbb{R}^4$ represents geometric constraints including spatial coordinates, compactness, and elongation measures derived from prototype feature statistics.

The geometric constraints are computed from prototype feature statistics to capture spatial and morphological properties:
\begin{align}
\text{spatial}_x &= \tanh\left(\frac{1}{D}\sum_{i=1}^{D} p_i\right) \times 0.5 \\
\text{spatial}_y &= \tanh\left(\sqrt{\frac{1}{D}\sum_{i=1}^{D} (p_i - \bar{p})^2}\right) \times 0.5 \\
\text{compactness} &= \exp\left(-\sqrt{\frac{1}{D}\sum_{i=1}^{D} (p_i - \bar{p})^2}\right) \\
\text{elongation} &= \frac{\max_i |p_i - \bar{p}|}{\frac{1}{D}\sum_{i=1}^{D} |p_i - \bar{p}| + \epsilon}
\end{align}

where $\mathbf{p} = [p_1, p_2, \ldots, p_D]$ is the prototype feature vector, $\bar{p}$ is the feature mean. While these constraints are derived from feature statistics, they serve as effective proxies for spatial characteristics of anatomical structures. The spatial coordinates $\text{spatial}_x$ and $\text{spatial}_y$ encode the feature centroid and variance respectively, providing positional information. The compactness measure captures feature concentration, with higher values indicating more concentrated features. The elongation ratio distinguishes compact from elongated structures by quantifying directional variance.

The complete geometric constraint vector is:
\begin{align}
\mathbf{g}_{geom} = [\text{spatial}_x, \text{spatial}_y, \text{compactness}, \text{elongation}]
\end{align}
To ensure balanced contribution from all constraints, each component is normalized to [0,1] range using min-max scaling before concatenation.

The spatial conditioning is integrated into the noise prediction process through adaptive injection that varies with the denoising timestep:
\begin{equation}
\epsilon_{enhanced} = \epsilon_\theta(\mathbf{p}_t, t) + \alpha_t \cdot \mathbf{c}_{spatial}
\end{equation}
where $\alpha_t = 0.1 \times (1 - t/T)$ provides stronger spatial guidance in early denoising steps and gradually reduces influence as the process approaches completion. This adaptive injection strategy provides strong spatial guidance during early recovery of coarse anatomical structures. 

\begin{figure}
    \centering
    \includegraphics[width=\columnwidth]{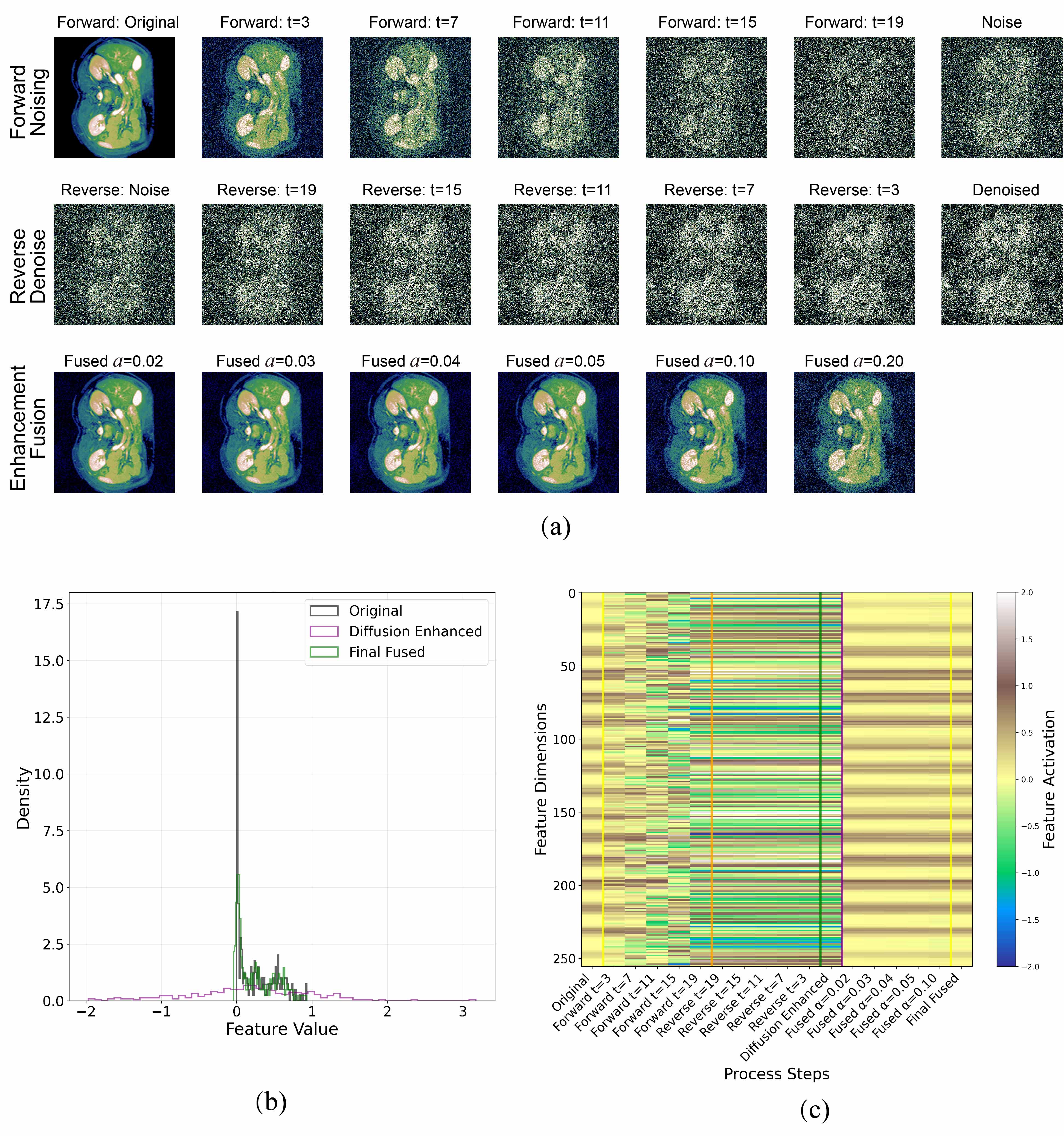}
    \caption{(a) Spatial-aware diffusion prototype heatmap evolution with forward, reverse diffusion and fusion process across timesteps (b) Density analysis of feature activation through diffusion and fusion. The original features exhibit high variance with noises and outliers, while the final fused features demonstrate improved smoothness while preserving essential activation patterns. (c) Feature evolution heatmap of diffusion process with multi-stage fusion.}
    \label{fig:spatial}
\end{figure}

\noindent\textbf{Conservative Enhancement Fusion.}
To preserve the reliability of original prototypes while incorporating diffusion benefits, we employ a conservative fusion strategy that adaptively combines original and enhanced prototypes:
\begin{align}
w_{\text{fidelity}} &= \sigma(\theta_{\text{fidelity}}), \quad w_{\text{diversity}} = \sigma(\theta_{\text{diversity}}) \\
\mathbf{p}_{\text{enhanced}} &= \frac{w_{\text{fidelity}} \mathbf{p}_0 + w_{\text{diversity}} \mathbf{p}_{\text{diff}}}{w_{\text{fidelity}} + w_{\text{diversity}}}
\end{align}
where $\mathbf{p}_0$ is the original prototype, $\mathbf{p}_{\text{diff}}$ is the diffusion-enhanced prototype, and $\theta_{\text{fidelity}}$, $\theta_{\text{diversity}}$ are learnable parameters controlling the fusion weights. This mechanism learns to combine the original prototype with the diverse diffusion-enhanced prototype to improve segmentation accuracy.


\subsection{Loss Function}

\noindent\textbf{Training-Inference Consistency.} 
DPL maintains consistency between training and inference by applying identical prototype enhancement and fusion mechanisms in both phases. During training, the segmentation loss $\mathcal{L}_{seg}$ optimizes based on predictions using enhanced prototypes $\mathbf{p}_{enhanced}$, while the diffusion loss $\mathcal{L}_{diffusion}$ provides regularization for learning robust prototype distributions. The same fusion weights $w_f$ and $w_d$ learned during training are applied during inference, ensuring consistent prototype representations. This unified approach eliminates trainin g-inference discrepancy while preserving the benefits of diffusion-based enhancement.

The DPL framework is optimized using a composite loss function that integrates segmentation, prototype alignment, and diffusion regularization objectives:
\begin{equation}
\mathcal{L}_{total} = \mathcal{L}_{seg} + \mathcal{L}_{align} + \beta \mathcal{L}_{diffusion}
\end{equation}
where each component serves a specific purpose in the learning process.

\noindent\textbf{Segmentation Loss ($\mathcal{L}_{seg}$).} The segmentation loss uses cross-entropy on predictions from enhanced prototypes for pixel-wise optimization:
\begin{equation}
\mathcal{L}_{seg} = -\frac{1}{HW} \sum_{h=1}^{H} \sum_{w=1}^{W} \sum_{c=1}^{C} y_{h,w,c} \log(\hat{y}_{h,w,c})
\end{equation}
Here, $y_{h,w,c}$ is the ground-truth label and $\hat{y}_{h,w,c}$ is the predicted probability for class $c$ at pixel $(h,w)$. The prediction $\hat{y}_{h,w,c}$ is calculated by matching query features against the enhanced prototype $\mathbf{p}_{\text{enhanced}}$.

\noindent\textbf{Alignment Loss ($\mathcal{L}_{align}$).} The alignment loss ensures consistency between enhanced prototypes from support and query features:
\begin{equation}
\mathcal{L}_{align} = \frac{1}{K} \sum_{k=1}^{K} \|P_k^{support,enhanced} - P_k^{query,enhanced}\|_2^2
\end{equation}
where $P_k^{support,enhanced}$ and $P_k^{query,enhanced}$ denote the $k$-th class enhanced prototypes from the support and query sets, both generated via the same diffusion and fusion pipeline.

\noindent\textbf{Diffusion Loss ($\mathcal{L}_{diffusion}$).} The diffusion regularization loss optimizes the denoising objective to learn the underlying prototype distribution:
\begin{equation}
\mathcal{L}_{diffusion} = \mathbb{E}_{P_0, \epsilon, t} \|\epsilon - \epsilon_\theta(P_t, t, S)\|_2^2
\end{equation}
where $P_0$ represents the clean prototype, $\epsilon \sim \mathcal{N}(0, I)$ is Gaussian noise, $P_t$ is the noisy prototype at timestep $t$, and $S$ denotes spatial conditioning information.
The weighting parameter $\beta = 0.02$ balances diffusion regularization with primary segmentation objectives, as validated in our ablation study (Table~\ref{tab:ablation}), ensuring prototype enhancement provides sufficient regularization without overwhelming core learning.

%% file: 4.experiment.tex
\section{Experiments}

\subsection{Datasets and Evaluation Metric}

For evaluating the proposed method, experiments are performed on two extensively utilized datasets: abdominal MR dataset (ABD-MRI)~\cite{ref17} and abdominal CT dataset (ABD-CT)~\cite{ref18}. The ABD-MRI dataset from the CHAOS challenge (ISBI 2019)~\cite{ref17} contains 20 3D MRI scans with annotations for left kidney, right kidney, liver, and spleen. The ABD-CT dataset from the Multi-Atlas Abdomen Labeling challenge (MICCAI 2015)~\cite{ref18} contains 30 3D abdominal CT scans from clinical patients with various pathologies.

Following standard few-shot learning protocols~\cite{ref12,ref13}, we enforce strict class separation where test classes are completely excluded from training images to ensure truly unseen evaluation. This setting provides the most rigorous assessment of one-shot segmentation performance.
Performance is measured using the Dice Similarity Coefficient (DSC), defined as:

\begin{equation}
\text{DSC}(M_p, M) = \frac{2|M_p \cap M|}{|M_p| + |M|} \times 100%
\end{equation}

where $M_p$ and $M$ represent predicted and ground truth segmentations. Each experiment is conducted for 3 epochs and the average DSC across all query classes is reported.

\subsection{Implementation Details}

The model is implemented in PyTorch and trained using SGD optimizer with momentum $0.9$ and weight decay $1 \times 10^{-4}$. The learning rate is set to $1 \times 10^{-3}$ for backbone parameters and $1 \times 10^{-7}$ for classification layers, with multi-step decay using gamma $0.95$. The model is trained for $30,000$ iterations under $1$-shot setting on a single RTX 4090 GPU.

\subsection{Comparison with State-of-the-Art Methods}

We compare DPL against recent few-shot segmentation methods including SE-Net \cite{senet2020}, prototype-based approaches (PANet \cite{ref5}, ALPNet \cite{alpnet2022}), query-guided methods (Q-Net \cite{qnet2023}, ADNet \cite{adnet2023}), and the latest self-refined prototype network DSPNet \cite{dspnet2024}. Tables~\ref{ABD-MRI} and~\ref{ABD-CT} present quantitative results on both ABD-MRI and ABD-CT datasets.

\noindent\textbf{ABD-MRI Dataset Performance.} On the ABD-MRI dataset, DPL achieves 82.60\% mean DSC, establishing new state-of-the-art performance by substantially outperforming DSPNet (79.29\%) by 3.31 percentage points, as shown in Table~\ref{ABD-MRI}. The improvement is particularly pronounced compared to traditional prototype-based approaches, with DPL surpassing PANet by 46.00 percentage points, highlighting the effectiveness of diffusion-enhanced prototypes. Organ-specific analysis reveals exceptional performance across all structures, achieving 91.45\% and 87.04\% DSC for right and left kidneys respectively, and 73.94\% for spleen segmentation.

\begin{center}
    \includegraphics[width=\columnwidth]{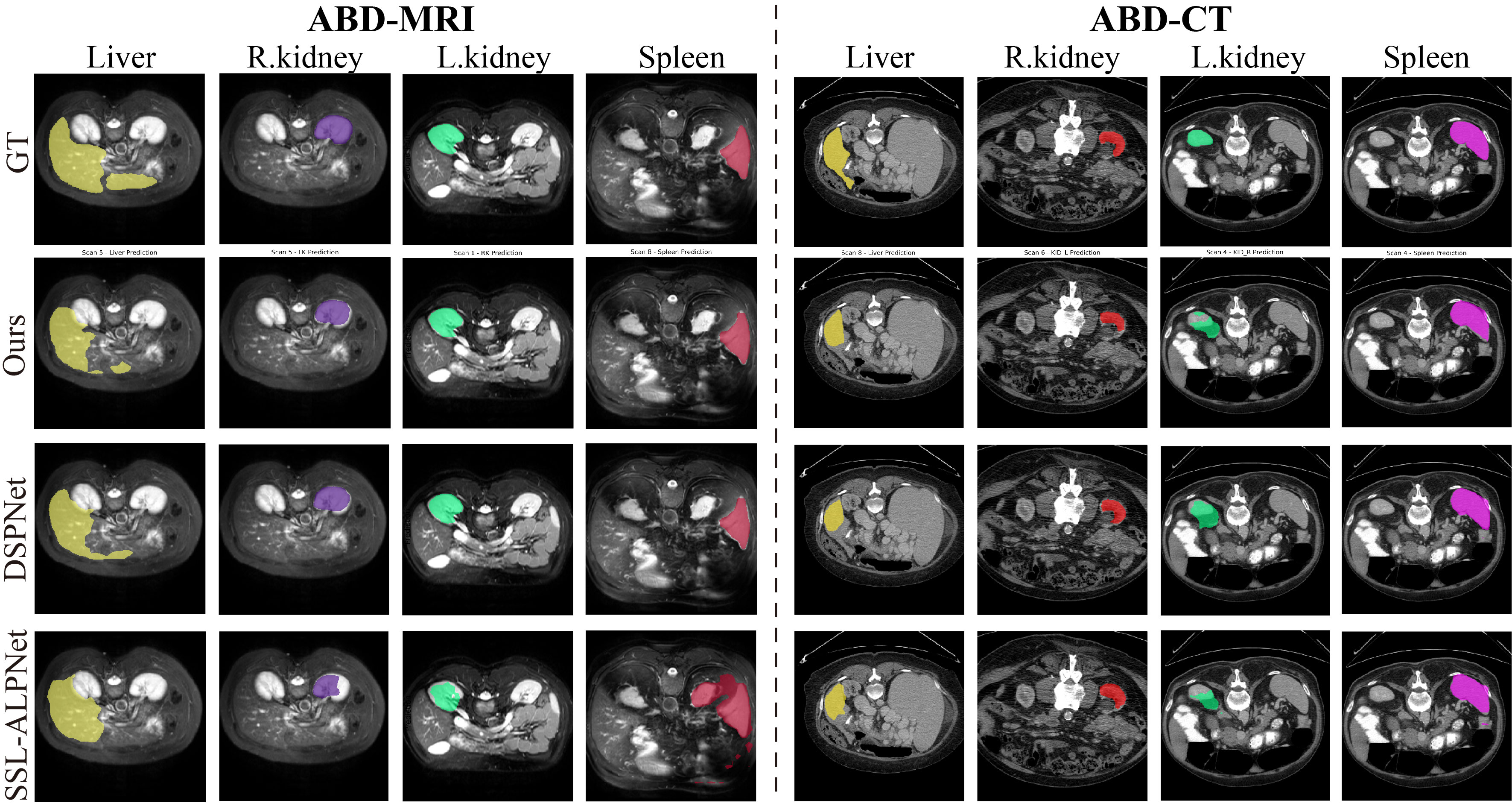}
    \captionof{figure}{Visualisation on ABD-MRI and ABD-CT. Top to Bottom: Ground-truth, DPL(ours) segmentation, DSPNet segmentation, and SSL-ALP segmentation results. }
    \label{fig:visual}
\end{center}

\begin{table}[H]
    \centering
    \caption{One-shot segmentation results on ABD-MRI dataset. All values are reported as Dice scores (\%). The best performance is marked in \textbf{bold}.}
    \label{ABD-MRI}
    \vspace{0.5em}
    \begin{tabular*}{\columnwidth}{l|*{5}{c}}
    \hline
    Method & Liver & R.kidney & L.kidney & Spleen & Avg\\
    \hline
    SE-Net (2020) \cite{senet2020} & 30.11 & 48.20 & 44.85 & 45.76 & 42.23\\
    ADNet (2023) \cite{adnet2023} & 69.55 & 78.80 & 80.30 & 63.20 & 72.96\\
    Q-Net (2023) \cite{qnet2023} & 72.20 & 83.10 & 76.80 & 70.70 & 75.20\\
    PANet (2019) \cite{ref5} & 48.02 & 32.15 & 36.40 & 29.88 & 36.61\\
    ALPNet (2022) \cite{alpnet2022} & 71.00 & 77.40 & 74.60 & 67.40 & 72.60\\
    DSPNet (2024) \cite{dspnet2024} & 74.33 & 87.88 & 85.62 & 69.31 & 79.29\\
    DPL(Ours) & \textbf{77.97} & \textbf{91.45} & \textbf{87.04} & \textbf{73.94} & \textbf{82.60}\\
    \hline
    \end{tabular*}
\end{table}

\noindent\textbf{ABD-CT Dataset Performance.} On the ABD-CT dataset, DPL achieves 76.33\% mean DSC, substantially outperforming the previous best method DSPNet (72.79\%) by 3.54 percentage points, as shown in Table~\ref{ABD-CT}. The approach demonstrates consistent improvements across all organs, with particularly notable gains in liver segmentation (77.78\% vs DSPNet's 69.32\%) and left kidney segmentation (80.23\% vs DSPNet's 78.01\%). The cross-modality validation between MRI and CT datasets confirms the robustness and generalizability of the diffusion-enhanced prototype approach.

\begin{table}[H]
    \centering
    \caption{One-shot segmentation results on ABD-CT dataset. All values are reported as Dice scores (\%). The best performance is marked in \textbf{bold}.}
    \label{ABD-CT}
    \vspace{0.5em}
    \begin{tabular*}{\columnwidth}{l|*{5}{c}}
    \hline
    Method & Liver & R.kidney & L.kidney & Spleen & Avg\\
    \hline
    SE-Net (2020) \cite{senet2020} & 34.05 & 13.20 & 25.38 & 42.10 & 28.68\\
    ADNet (2023) \cite{adnet2023} & 66.80 & 71.90 & 74.60 & 69.01 & 70.58\\
    Q-Net (2023) \cite{qnet2023} & 67.10 & 54.20 & 68.30 & 58.20 & 61.95\\
    PANet (2019) \cite{ref5} & 59.12 & 51.80 & 54.67 & 54.41 & 55.00\\
    ALPNet (2022) \cite{alpnet2022} & 65.80 & 48.40 & 68.10 & 65.20 & 61.88\\
    DSPNet (2024) \cite{dspnet2024} & 69.32 & 74.54 & 78.01 & 69.31 & 72.79\\
    DPL(Ours) & \textbf{77.78} & \textbf{76.30} & \textbf{80.23} & \textbf{71.00} & \textbf{76.33}\\
    \hline
    \end{tabular*}
\end{table}

\subsection{Ablation Studies}

\noindent\textbf{Module Ablation Analysis.} 
The ablation study in Table~\ref{tab:ablation} validates each framework component. Removing forward diffusion most significantly reduced performance (66.88\%), while removing reverse diffusion (69.92\%) and spatial injection (71.42\%) showed moderate impacts. Removing fusion achieved competitive performance (71.90\%), indicating core diffusion processes provide the primary benefits. The complete DPL framework achieved optimal performance at 75.12\%.

\begin{center}
\footnotesize
\captionof{table}{Ablation study results on multi-organ segmentation task. All values are reported as Dice scores (\%). }
\label{tab:ablation}
\begin{tabular}{>{\centering\arraybackslash}m{1.5cm}|>{\centering\arraybackslash}m{1cm}|>{\centering\arraybackslash}m{1cm}|>{\centering\arraybackslash}m{0.7cm}|>{\centering\arraybackslash}m{1cm}|>{\centering\arraybackslash}m{1.0cm}}
\hline
Method & Forward & Reverse & Spatial & Fusion & Avg\\
\hline
Baseline & $\times$ & $\times$ & $\times$ & $\times$ & 69.19\\
\hline
No Forward Diffusion & $\times$ & T=20, $\alpha$=0.03 & $\checkmark$ & $w_f$=0.8, $w_d$=0.2 & 66.88\\
\hline
No Reverse Diffusion & T=20, $\alpha$=0.03 & $\times$ & $\times$ & $w_f$=0.8, $w_d$=0.2 & 69.92\\
\hline
No Spatial Injection & T=20, $\alpha$=0.03 & T=20, $\alpha$=0.03 & $\times$ & $w_f$=0.8, $w_d$=0.2 & 71.42\\
\hline
No Fusion & T=20, $\alpha$=0.03 & T=20, $\alpha$=0.03 & $\checkmark$ & $\times$ & 71.90\\
\hline
DPL-Full & T=20, $\alpha$=0.03 & T=20, $\alpha$=0.03 & $\checkmark$ & $w_f$=0.8, $w_d$=0.2 & \textbf{75.12}\\
\hline
\end{tabular}
\end{center}

\noindent\textbf{Hyperparameter Sensitivity Analysis.}
Table~\ref{tab:sensitivity} demonstrates the sensitivity of DPL to key hyperparameters. The analysis shows that $\beta=0.02$ provides optimal balance, with larger values (0.05, 0.1) leading to over-regularization. The spatial conditioning strength $\alpha_t=0.1$ achieves best performance, while extreme values (0.05, 0.2) show degradation. Timesteps $T=20$ proves sufficient, with $T=50$ showing marginal gains at higher computational cost.

\begin{table}[H]
\centering
\caption{Hyperparameter sensitivity analysis showing the impact of key parameters on segmentation performance. Best performance metrics are highlighted in \textbf{bold}.}
\fontsize{8pt}{10pt}\selectfont
\label{tab:sensitivity}
\begin{tabular}{>{\centering\arraybackslash}p{0.05cm}|>{\centering\arraybackslash}p{0.3cm}>{\centering\arraybackslash}p{0.3cm}>{\centering\arraybackslash}p{0.3cm}|>{\centering\arraybackslash}p{0.7cm}>{\centering\arraybackslash}p{0.7cm}>{\centering\arraybackslash}p{0.7cm}>{\centering\arraybackslash}p{0.7cm}>{\centering\arraybackslash}p{0.7cm}}
\hline
\multirow{2}{*}{\#} & \multicolumn{3}{c|}{Diffusion Params} & \multicolumn{5}{c}{Performance (Dice \%)} \\
\cline{2-4} \cline{5-9}
 & $\beta$ & $\alpha_t$ & $T$ & Liver & R.Kid & L.Kid & Spleen & Mean \\
\hline
1 & 0.01 & 0.1 & 20 & 72.15 & 83.42 & 81.23 & 58.94 & 73.94 \\
2 & \textbf{0.02} & \textbf{0.1} & \textbf{20} & \textbf{73.31} & \textbf{84.23} & \textbf{82.37} & \textbf{60.55} & \textbf{75.12} \\
3 & 0.05 & 0.1 & 20 & 72.88 & 83.76 & 81.89 & 59.47 & 74.50 \\
4 & 0.1 & 0.1 & 20 & 71.42 & 82.31 & 80.65 & 57.83 & 73.05 \\
5 & 0.02 & 0.05 & 20 & 72.45 & 83.18 & 81.67 & 59.12 & 74.11 \\
6 & 0.02 & 0.2 & 20 & 72.89 & 83.91 & 82.01 & 59.78 & 74.65 \\
7 & 0.02 & 0.1 & 10 & 71.98 & 82.67 & 80.95 & 58.43 & 73.51 \\
8 & 0.02 & 0.1 & 50 & 72.76 & 83.54 & 81.78 & 59.31 & 74.35 \\
\hline
\end{tabular}
\end{table}

\section{Conclusion}

This work introduces Diffusion Prototype Learning (DPL), a novel framework that reformulates prototype construction in one-shot medical image segmentation through diffusion-based feature space exploration. By modeling prototypes as learnable probability distributions rather than deterministic point estimates, DPL addresses the critical limitation of traditional averaging operations in capturing anatomical variability. Extensive experiments on abdominal MRI and CT datasets demonstrate substantial improvements, establishing new state-of-the-art benchmarks. Systematic ablation studies validate the complementary contributions of diffusion processes. 